\documentclass[letterpaper]{article} 
\usepackage{aaai24}  
\usepackage{times}  
\usepackage{helvet}  
\usepackage{courier}  
\usepackage[hyphens]{url}  
\usepackage{graphicx} 
\usepackage{natbib}  
\usepackage{caption} 
\usepackage{algorithm}
\usepackage{algorithmic}
\usepackage{subfigure}
\usepackage{amsmath}
\usepackage{amssymb}
\usepackage{multirow, makecell}
\usepackage{booktabs}
\usepackage{newfloat}
\usepackage{listings}
\DeclareCaptionStyle{ruled}{labelfont=normalfont,labelsep=colon,strut=off} 
\floatstyle{ruled}
\newfloat{listing}{tb}{lst}{}
\floatname{listing}{Listing}
\title{EarthVQA: Towards Queryable Earth via Relational Reasoning-Based Remote Sensing Visual Question Answering}
\author{
    Junjue Wang\textsuperscript{\rm 1}, Zhuo Zheng\textsuperscript{\rm 2}, Zihang Chen\textsuperscript{\rm 1}, Ailong Ma\textsuperscript{\rm 1}, Yanfei Zhong\textsuperscript{\rm 1}\thanks{Corresponding author}
}
\affiliations{
    \textsuperscript{\rm 1}Wuhan University, \textsuperscript{\rm 2}Stanford University

    \{kingdrone, chenzihang, maailong007, zhongyanfei\}@whu.edu.cn, zhuozheng@cs.stanford.edu
}

\usepackage{bibentry}

\begin{document}

\maketitle

\begin{abstract}
    Earth vision research typically focuses on extracting geospatial object locations and categories but
    neglects the exploration of relations between objects and comprehensive reasoning. 
    Based on city planning needs, we develop a multi-modal multi-task VQA dataset (\textbf{EarthVQA}) to advance relational reasoning-based judging, counting, and comprehensive analysis.
    The EarthVQA dataset contains 6000 images, corresponding semantic masks, and 208,593 QA pairs with urban and rural governance requirements embedded.
    As objects are the basis for complex relational reasoning, we propose a Semantic OBject Awareness framework (\textbf{SOBA}) to advance VQA in an object-centric way.
    To preserve refined spatial locations and semantics, SOBA leverages a segmentation network for object semantics generation.
    The object-guided attention aggregates object interior features via pseudo masks, and bidirectional cross-attention further models
    object external relations hierarchically.
    To optimize object counting, we propose a numerical difference loss that dynamically adds difference penalties, unifying the classification and regression tasks.
    Experimental results show that SOBA outperforms both advanced general and remote sensing methods.
    We believe this dataset and framework provide a strong benchmark for Earth vision's complex analysis. The project page is at https://Junjue-Wang.github.io/homepage/EarthVQA.
\end{abstract}

\section{Introduction}
High-spatial resolution (HSR) remote sensing images can assist us in quickly obtaining essential information \cite{zvonkov2023openmapflow, xiao2023d2u}.
Most research focuses on the perception of object
categories and locations, deriving related tasks such as semantic segmentation \cite{RN787}, species detection \cite{ZHAO2022328}, and urban understanding \cite{shi2023multi}.
However, the existing methods and datasets ignore the relations between the geospatial objects, 
thus limiting their ability to knowledge reasoning in complex scenarios.
Especially in city planning \cite{bai2014society}, the relations between the transportation hubs and schools,
water situations around the farmland, and greenery distributions in residential areas
are also significant and urgent to be analyzed.
Hence, it is necessary to go beyond object perception and explore object relations,
bridging the gap between information and comprehensive knowledge \cite{li2022searching}.

\begin{figure*}[hbt]
    \centering
    \includegraphics[width=1\linewidth]{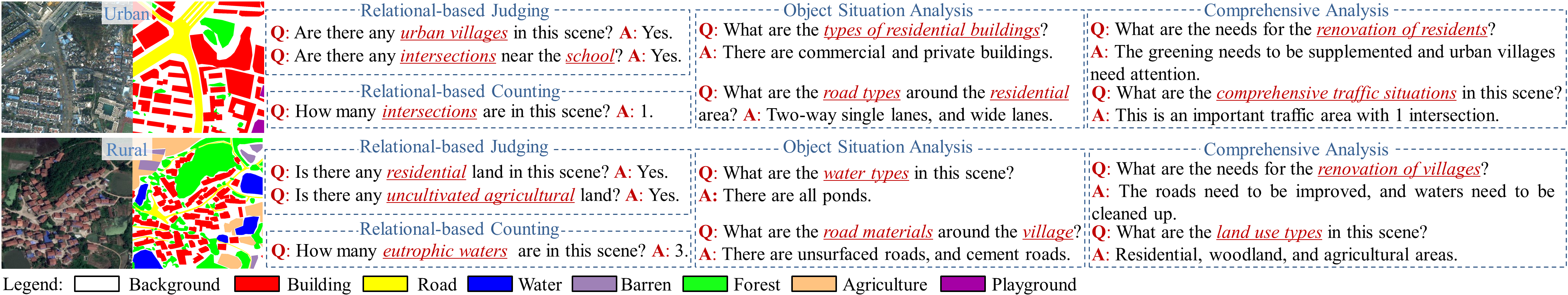}
    \caption{Urban and rural samples (image-mask-QA pairs) from the EarthVQA dataset. The QA pairs are designed to based on city planning needs, including judging, counting, object situation analysis, and comprehensive analysis types. 
    This multi-modal and multi-task dataset poses new challenges, requiring object-relational reasoning and knowledge summarization.} 
    \label{fig:dataset}
\end{figure*}

Visual question answering (VQA) aims to answer customized questions by searching for visual clues in the provided image.
Since linguistic questions determine the task properties, the algorithms are flexible and can be developed for reasoning required answers.
Recently, preliminary VQA datasets and methods have emerged in the remote sensing field \cite{9088993, zheng2021mutual, rahnemoonfar2021floodnet}.
However, most of these researches have the following drawbacks:
1) As for most datasets, QA pairs are automatically labelled based on existing data, such as Open Street Map (OSM) and classification datasets. 
Most tasks are simple counting and judging questions with no relational reasoning required.
The automatic QA pairs do not match actual needs, limiting their practicalities.
2) The development of the remote sensing VQA model lags, and most research directly fuses the global visual and language features to predict the final answers.
They ignore the local semantics and relations, which are unsuitable for the complex reasoning of multiple geospatial objects.
To this end, we propose a multi-modal multi-task VQA dataset and a semantic object awareness framework to advance complex remote sensing VQA tasks.
Main contributions are as follows:

\begin{itemize}
  \item [1)] 
  We propose the EarthVQA dataset with triplet samples (image-mask-QA pairs).
  The 208,593 QA pairs encompass six main categories\footnote{basic judging, basic counting, relational-based judging, relational-based counting, object situation analysis, and comprehensive analysis.}.
  EarthVQA features diverse tasks from easy basic judging to complex relation reasoning and even more challenging comprehensive analysis.
  Specifically, the residential environments, traffic situations, and renovation needs of waters and unsurfaced roads are explicitly embedded in various questions.
  \item [2)]
  To achieve relational reasoning-based VQA, we propose a semantic object awareness framework (SOBA).
  SOBA utilizes segmentation visual prompts and pseudo masks to generae pixel-level features with accurate locations.
  The object awareness-based hybrid attention models the relations for object-guided semantics and
  bidirectionally aggregates multi-modal features for answering.
  \item [3)]
  To add distance sensitivity for regression questions, we propose a numerical difference (ND) loss.
  The dynamic ND penalty is seamlessly integrated into cross-entropy loss for the regression task.
  ND loss introduces the sensitivity of numerical differences into the model training.
\end{itemize}

\section{Related Work}
\label{sec:2}
\noindent \textbf{General visual question answering.}
The vanilla VQA model \cite{antol2015vqa} includes three parts: a convolutional neural network (CNN), a long-short term memory (LSTM), and a fusion classifier.
Specifically, CNN extracts visual features for input images, and LSTM embeds the language features for the questions.
Global features are interacted in the fusion classifier and finally generate the answer.
Based on this architecture, more powerful encoders and fusion modules were proposed.
To obtain local visual features, the bottom-up top-down attention (BUTD) mechanism \cite{anderson2018bottom} introduced objectness features generated by Faster-RCNN \cite{ren2015faster} pretrained on Visual Genome \cite{krishna2017visual} data.
For computational efficiency, a recurrent memory, attention, and composition (MAC) cell \cite{mac} was designed to explicitly model the relations between image and language features.
Similarly, the stacked attention network (SAN) \cite{yang2016stacked} located the relevant visual clues guided by question layer-by-layer.
By combining objectness features with attention, the modular co-attention network (MCAN) \cite{yu2019deep} adopted a transformer to model
intra- and inter-modality interactions.
To alleviate language biases, D-VQA \cite{wen2021debiased} applied an unimodal bias detection module to
explicitly remove negative biases.
BLIP-2 \cite{blip2} and
Instruct-BLIP \cite{dai2023instructblip} bridge the large pre-trained vision and language models using the Q-Former, addressing VQA as a generative task. 
Besides, many advanced VQA methods \cite{marino2021krisp} eliminate statistical bias by accessing external databases.

\noindent \textbf{Remote sensing visual question answering.}
The remote sensing community has some early explorations including both datasets and methods.
The QA pairs of the RSVQA dataset \cite{9088993} are queried from OSM, and images are obtained from Sentinel-2 and other sensors.
RSIVQA dataset \cite{zheng2021mutual} is automatically generated from 
the existing classification and object detection datasets, \textsl{i.e.}, AID \cite{xia2017aid}, HRRSD \cite{zhang2019hierarchical}, etc.
The FloodNet \cite{rahnemoonfar2021floodnet} dataset was designed for disaster assessment, mainly concerned with the inundation of roads and buildings.

Compared with these datasets, the EarthVQA dataset has two advantages: 
\textbf{1) Multi-level annotations.} The annotations include pixel-level semantic labels, object-level analysis questions, and scene-level land use types.
 Supervision from different perspectives advances a comprehensive understanding of complex scenes.
\textbf{2) Complex and practical questions.} The existing datasets focus on counting and judging questions, 
which only involve simple relational reasoning about one or two types of objects.
In addition to counting and judging, EarthVQA also contains various object analysis and comprehensive analysis questions.
These promote complex relational reasoning by introducing spatial or semantic analysis of more than three types of objects.
Only basic judging
and counting answers are auto-generated from the LoveDA
masks. Other reasoning answers (Figure~\ref{fig:dataset}) are manually annotated (reasoning distances, layouts,
topologies, sub-properties, etc) for city planning needs.

Remote sensing algorithms are mainly modified from general methods, for example, RSVQA is based on vanilla VQA \cite{antol2015vqa}.
RSIVQA \cite{zheng2021mutual} designed a mutual attention component to improves interactions for multi-modal features.
CDVQA \cite{yuan2022change} introduced VQA into change detection task.
We novelly introduce pixel-level prompts for the guidance of VQA tasks, making it suitable for scenes with compact objects.

\begin{figure*}[!hbt]
    \subfigure[Annotation procedure of relational reasoning-based QA.]{
      \includegraphics[width=0.31\linewidth]{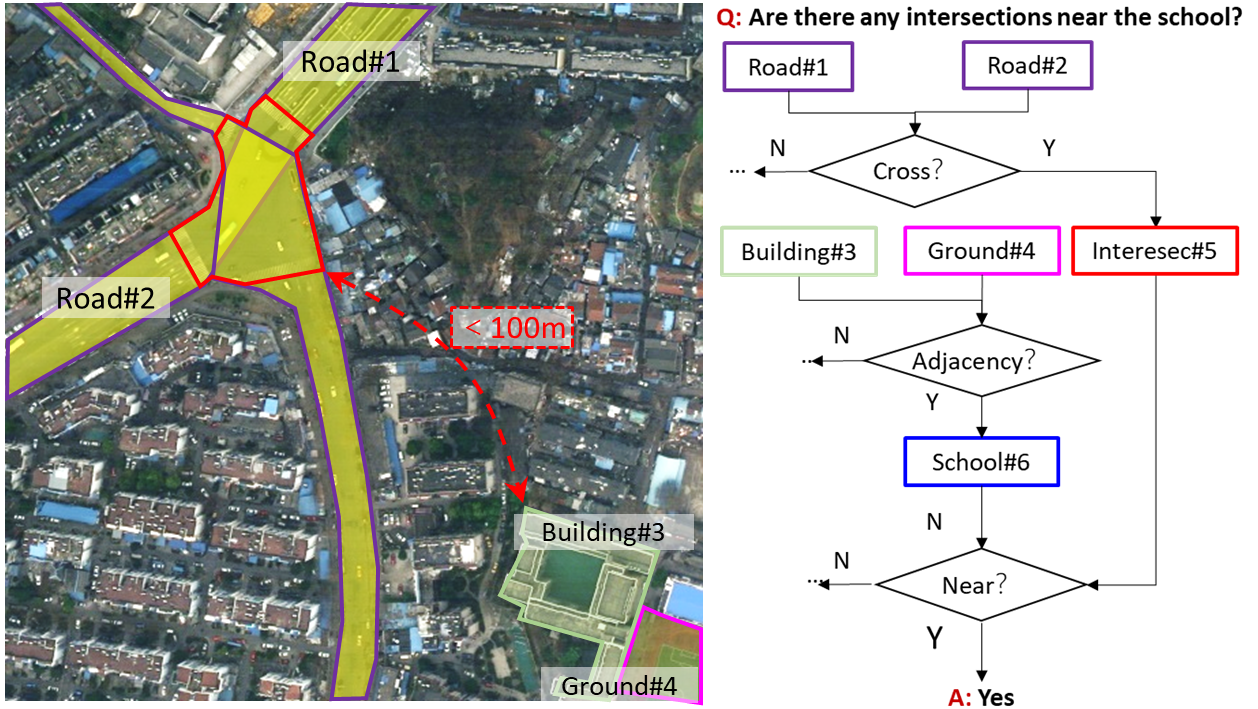}
      \label{fig:urban_ann_proc}
      }
    \subfigure[Statistics of questions.]{
      \includegraphics[width=0.26\linewidth]{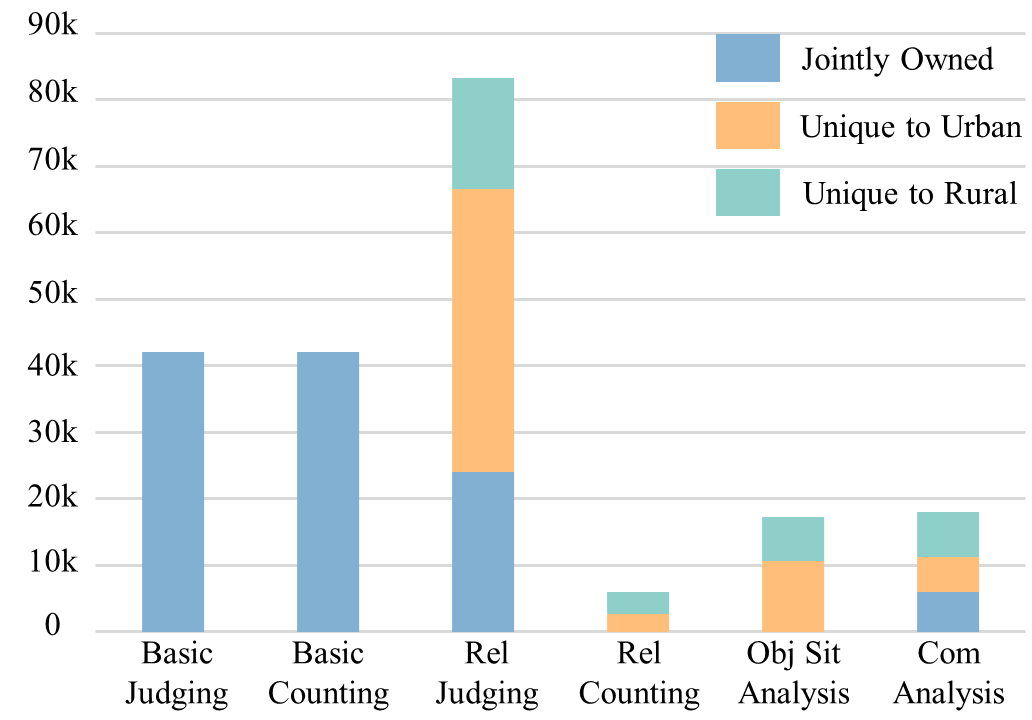}
      \label{fig:questions_statics}
      }
      \subfigure[Distributions of the top 30 most frequent answers.]{
      \includegraphics[width=0.38\linewidth]{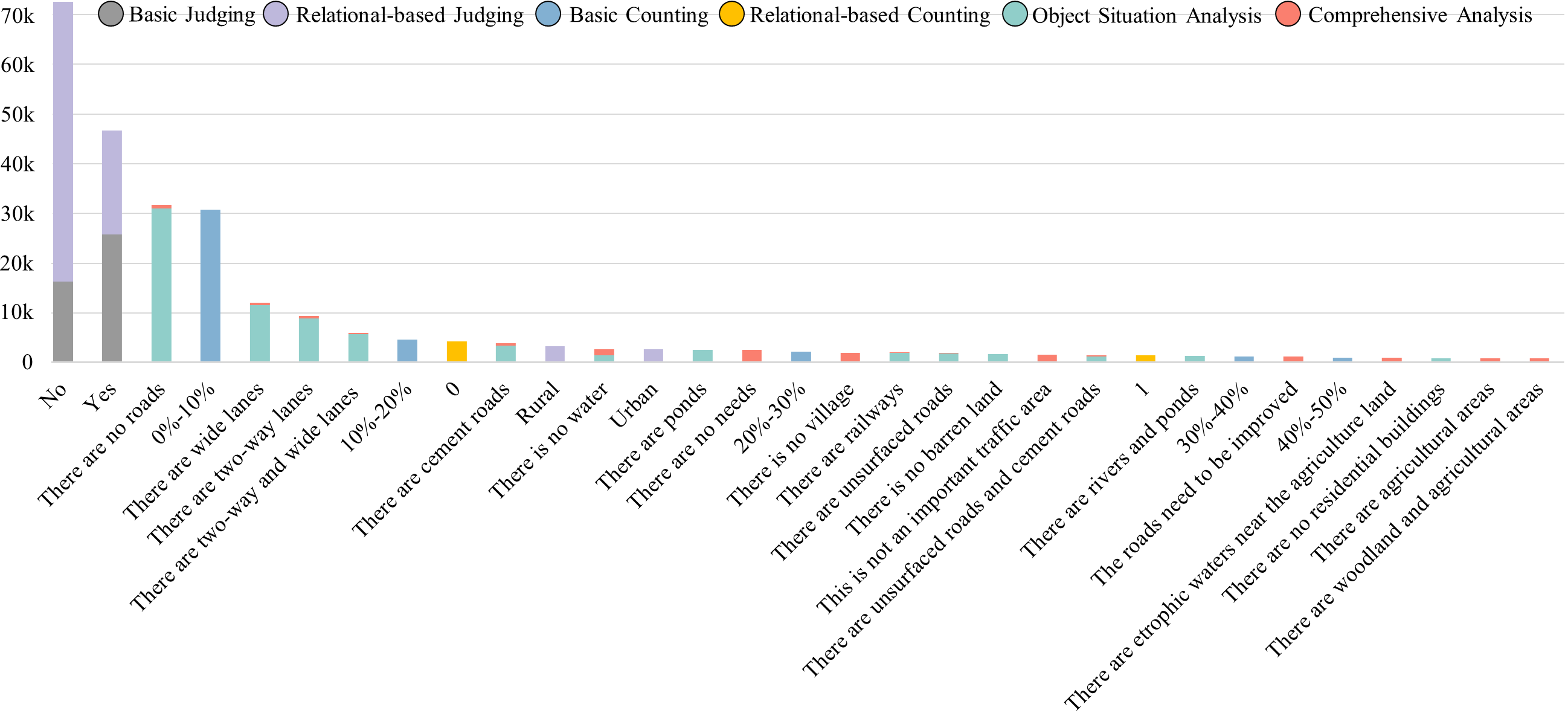}
      \label{fig:analysis_ans}
      }
    \caption{Details of questions and answers in EarthVQA dataset.
    Each urban image has a set of 42 questions and each rural image has a set of 29 questions, ensuring relatively balanced for each question. 
    The imbalanced distributions of answers bring more challenges when faced with the actual Earth environment.
    }
    \label{fig:statics}
  \end{figure*}

\section{EarthVQA Dataset}
\label{sec:3}

The EarthVQA dataset was extended from the LoveDA dataset \cite{wang2021loveda}, which encompasses
18 urban and rural regions from Nanjing, Changzhou, and Wuhan. 
LoveDA dataset provides 5987 HSR images and semantic masks with seven common land-cover types.
There are three significant revisions: \textsl{1) Quantity expansion.} 8 urban and 5 rural samples are added to expand capacity to 6000 images (WorldView-3 0.3m).
\textsl{2) Label refinement.} `playground' class was added as an important artificial facility, and some errors were revised for semantic labels.
\textsl{3) Addition of QA pairs.} We added 208,593 QA pairs to introduce VQA tasks for city planning.
Each urban image has 42 QAs and each rural image has 29 QAs.  
Following the balanced division \cite{wang2021loveda},
\texttt{train} set includes 2522 images with 88166 QAs, \texttt{val} set includes 1669
images with 57202 QAs, and
\texttt{test} set includes 1809 images with 63225 QAs.

\noindent \textbf{Annotation procedure.}
EarthVQA currently does not involve ambiguous
questions such as geographical orientations.
As for `Are there any intersections near the school?' in Figure~\ref{fig:urban_ann_proc},
by judging the topology, the recognized Road\#1 and Road\#2 firstly form Intersection\#5.
Similarly, Ground\#4 and Building\#3 jointly form the scene of School\#6.
We use the ArcGIS toolbox to calculate the polygon-to-polygon distance between School\#6 and Intersection\#5, and obtain 94.8m $<$ 100m.
Hence, the final answer is `Yes'.
Each step has fixed
thresholds and conditions.

\noindent \textbf{Statistics for questions.}
As is shown in Figure~\ref{fig:questions_statics},
urban and rural scenes have common and unique questions according to the city planning demands.
The number of questions for urban and rural is balanced, eliminating geographical statistical bias.
Basic questions involve the statistics and inference of a certain type of object, \textsl{i.e.}, `What is the area of the forest?'.
Relational-based questions require semantic or spatial relational reasoning between different objects.
Comprehensive analysis focuses on more than three types of objects, including a summarization of traffic facilities,
water sources around agriculture, land-use analysis, etc.

\noindent \textbf{Statistics for answers.}
As shown in Figure~\ref{fig:analysis_ans},
we selected
the top 30 most frequent answers from 166 unique answers in the dataset.
Similar to the common VQA datasets, the imbalanced distributions
of answers bring more challenges when faced with
the actual Earth environment.
\section{Semantic object awareness framework} 
\label{sec:4}
To achieve efficient relational reasoning,
we design the SOBA framework for complex city scenes.
SOBA includes a two-stage training: 1) semantic segmentation network training for generating visual prompts and pseudo masks;
and 2) hybrid attention training for reasoning and answering.

\begin{figure*}[hbt]
  \centering
  \includegraphics[width=1\linewidth]{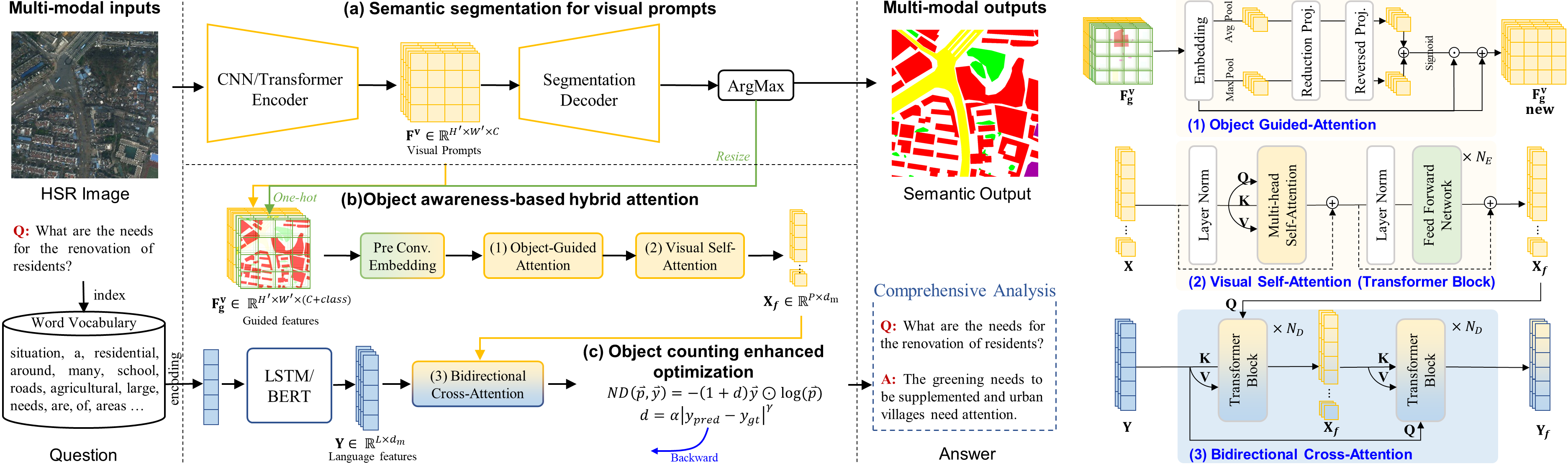}
  \caption{\textbf{(Left)} The architecture of SOBA includes (a) deep semantic segmentation for visual prompts; (b) object awareness-based hybrid attention (\textbf{Right} shows the details); and (c) object counting enhanced optimization.
  } 
  \label{fig:framework}
\end{figure*}

\subsection{Semantic segmentation for visual prompts}
Faced with HSR scenes containing multiple objects, we novelly adopt a segmentation network for refined guidance.
For an input image $\mathbf{I} \in \mathbb{R}^{H \times W \times 3}$, we utilize the encoder outputs $\mathbf{F}^v \in \mathbb{R}^{H' \times W' \times C}$ as the visual prompts.
$C$ denotes feature dimension and $H' = \frac{H}{32}, W'=\frac{W}{32}$ according to common settings.
Pseudo semantic output $\mathbf{M}^v \in \mathbb{R}^{H \times W}$ is also adopted for object awareness.
Compared with the existing Faster-RCNN based algorithms \cite{yu2019deep, anderson2018bottom} which averages box features in one vector, the pixel-level visual prompts preserve the locations and semantic details inside objects.
This contributes to the modeling of various compact objects in HSR scenes.

\subsection{Object awareness-based hybrid attention}
Guided by questions and object masks,
Object awareness-based hybrid attention reasons
visual cues for final answers.
As is shown in Figure~\ref{fig:framework},
there are three components: 
1) object-guided attention (OGA), 2) visual self-attention (VSA), and 3) bidirectional cross-attention (BCA).

\noindent \textbf{OGA for object aggregation.}
Because segmentation output has object details $\mathbf{M}^v$ (including categories and boundaries),
it is adopted to explicitly enhance visual prompts.
OGA is proposed to dynamically weight $\mathbf{F}^v$ and $\mathbf{M}^v$ from the channel dimension.
Using the nearest interpolation,
$\mathbf{M}^v$ is firstly resized into the same size as $\mathbf{F}^v$.
One-hot encoding followed with a pre-convolutional embedding then serializes the object semantics. The embedding contains a 3 $\times$ 3 convolution, a batch normalization, and a ReLU. 
They are concatenated to obtain object-guided features $\mathbf{F}^v_g$ as inputs for OGA.
The reduction and reverse projections further refine the features dimensionally.
After activation, we use 
the refined features to calibrate subspaces of $\mathbf{F}^v_g$ from the channel dimension.

\noindent \textbf{VSA for feature enhancement.}
To capture long-distance relations between geospatial objects, VSA \cite{dosovitskiy2020vit} hierarchically transforms the refined features.  
VSA includes $N_e$ transformer blocks, and each includes 
a multi-head self-attention (MSA) and a feed-forward network (FFN).
The refined features are reduced by a $1 \times 1$ convolution and
reshaped to generate patches $\mathbf{X} \in \mathbb{R}^{P \times d_m}$.
$P = \frac{H}{32} \times \frac{W}{32}$ denotes token size and $d_m$ is hidden size.

At each block $i$, features are transformed into a triplet: $\mathbf{Q} = \mathbf{X}^{i-1}\mathbf{W}^q, \mathbf{K} = \mathbf{X}^{i-1}\mathbf{W}^k, \mathbf{V} = \mathbf{X}^{i-1}\mathbf{W}^v$,
where $\mathbf{W}^q$, $\mathbf{W}^k$, $\mathbf{W}^v \in \mathbb{R}^{d_m \times d_v}$ denote the weights of three linear projections and $d_v = d_m / M$ is the reduction dim of each head.
The self-attention firstly calculates the similarities between each patch and then weight their values: 
$Att(\mathbf{Q}, \mathbf{K}, \mathbf{V}) = softmax(\frac{\mathbf{Q}\mathbf{K}^{T}}{\sqrt{d_v}})\mathbf{V}$.
MSA repeats the attention operation $M$ times in parallel and concatenates outputs.
Finally, outputs are fused by a linear projection. Formally,
$MSA(\mathbf{Q}, \mathbf{K}, \mathbf{V}) = Concat(h_1, ..., h_M)\mathbf{W}^O$, where
$h_i = Att(\mathbf{Q}_i, \mathbf{K}_i, \mathbf{V}_i)$ and
$\mathbf{W}^O \in \mathbb{R}^{Md_v \times d_m}$ denotes projection weights.
MSA models long-distance dependency by calculating the similarities between each geospatial object.
FFN consists of two linear transformation layers, and a GELU to improve visual representations.
The formulation is shown as 
$FFN(\mathbf{X}^{i-1}) = GELU(\mathbf{X}^{i-1}\mathbf{W}_1)\mathbf{W}_2$,
where $\mathbf{W}_1 \in \mathbb{R}^{d_m \times d_f}, \mathbf{W}_2 \in \mathbb{R}^{d_f \times d_m}$ represent the learnable projection parameters.
$d_f$ denotes the hidden size of FFN.

\noindent \textbf{BCA for multi-modal interaction.}
BCA advances the interaction with visual and language features via a bidirectional fusion mechanism.
BCA consists of two series of $N_d$ transformer blocks. 
The first stage aggregates useful language features to enhance visual features $\mathbf{X}_f$
and second stage implicitly models object external relations according to keywords, boosting language features $\mathbf{Y}_f$.
The implementation can be formulated as follows:
\begin{equation}
  \begin{aligned}
    \mathbf{Q}_\texttt{V} &= \mathbf{X}\mathbf{W}^q, \mathbf{K}_\texttt{L} = \mathbf{Y}\mathbf{W}^k, \mathbf{V}_\texttt{L} = \mathbf{Y}\mathbf{W}^v \\
    \mathbf{X}_{f} &= \text{Att}(\mathbf{Q}_\texttt{V}, \mathbf{K}_\texttt{L}, \mathbf{V}_\texttt{L}) \\
    \mathbf{Q}_\texttt{L} &= \mathbf{Y}\mathbf{W}^q, \mathbf{K}_\texttt{V} = \mathbf{X}_{f}\mathbf{W}^k, \mathbf{V}_\texttt{V} = \mathbf{X}_{f}\mathbf{W}^v \\
    \mathbf{Y}_{f} &= \text{Att}(\mathbf{Q}_\texttt{L}, \mathbf{K}_\texttt{V}, \mathbf{V}_\texttt{V})
  \end{aligned}
  \label{eq:bca}
\end{equation}
Finally, the fused $\mathbf{X}_f$ and $\mathbf{Y}_f$ are used for the final analysis. 
Compared with previous research \cite{cascante2022simvqa} which only uses one-way cross-attention, bidirectional attention mechanism 
hierarchically aggregates multi-modal features 
by simulating the human process of finding visual cues \cite{savage2019ai}.
Besides,
we have also conducted comparative experiments with alternative
cross-attention variants in Table~\ref{tab:oneway_vs_bca} and Table~\ref{tab:bca}.
\subsection{Object counting enhanced optimization}
VQA tasks include both classification and regression (object counting) questions. However, existing methods regard them as a multi-classification task, which is processed with cross-entropy (CE) loss.
Eq.~\eqref{eq:ce} represents that CE loss is insensitive to the distance between predicted value and true value, and is therefore not suitable for the regression task.
\begin{equation}
  CE(\vec{p} ,\vec{y})  = -\vec{y} \odot log(\vec{p}) =\sum^{class}_{i=1} -y_i log(p_i) \label{eq:ce}
\end{equation}
where $\vec{y}$ specifies one-hot encoded ground truth and
$\vec{p}$ denotes predicted probabilities. 
To introduce difference penalty for the regression task, we add a modulating factor $d = \alpha |\mathbf{y}_{diff}|^{\gamma} =\alpha |\mathbf{y}_{pr} - \mathbf{y}_{gt}|^{\gamma}$
to CE loss. $\mathbf{y}_{pr}$ and $\mathbf{y}_{gt}$ represent the predicted and ground truth number, respectively. $\alpha \geq 0$ and $\gamma \geq 0$ are tunable distance awareness factors.
$d$ represents the distance penalty $d \propto \mathbf{y}_{diff}$. 
Finally, we design the numerical difference (ND) loss as follows:
\begin{equation}
  \begin{split}
  ND(\vec{p} ,\vec{y})  &= -(1 + d) \vec{y} \odot log(\vec{p}) \\
  &= -(1 + \alpha |\mathbf{y}_{diff}|^{\gamma}) \vec{y} \odot log(\vec{p}) \\
  &=  -(1 + \alpha |\mathbf{y}_{pr} - \mathbf{y}_{gt}|^{\gamma}) \sum^{class}_{i=1} y_i log(p_i)
\end{split}\label{eq:nl}
\end{equation}

ND loss unifies classification and regression objectives into one optimization framework.
$\alpha$ controls the overall penalty for regression tasks compared to classification tasks.
$\gamma$ determines the sensitivity of regression penalty to numerical differences.
As the $\alpha$ increases, the overall penalty increases, meaning that optimization focuses more
on regression tasks.
With $\alpha=0$, the ND loss degenerates into the original CE loss and the penalty is constant ($d = 0$ when $|\mathbf{y}_{diff}| \in [0, +\infty)$).
The sensitivity of the regression penalty increases as $\gamma$ increases,
and when $\gamma > 1$, the penalty curve changes from concave to convex.

\begin{table*}[!hbt]
  \caption{Compared results with other VQA methods on EarthVQA$\rm ^{\textsl{test}}$}
  \label{tab:comp_rescuevqa}
  \resizebox{1.0\linewidth}{!}{
  \begin{tabular}{l|c|ccccccc|ccc|cc}
  \multirow{2}{*}{Method}         & \multirow{2}{*}{Promp.}  & \multicolumn{6}{c}{$\uparrow$Accuracy(\%)}                                                                                          & \multirow{2}{*}{$\uparrow$OA(\%)}  & \multicolumn{2}{c}{$\downarrow$RMSE}                    & \multirow{2}{*}{$\downarrow$OR}  & Param. & FLOPs   \\ 
       &                & Bas Ju       & Rel Ju         & Bas Co      & Rel Co        & Obj An   & Com An         &             & Bas Co         & Rel Co        &               & (M)        & (B)                \\ \Xhline{1.0pt}
  \multicolumn{2}{l|}{$\star$\textsl{General methods}} & & & & & & & & & &\\
  SAN      & $\times$           & 87.59	& 81.79	& 76.26	& 59.23	& 55.00	& 43.25	& 75.66                & 1.1367	&1.3180	&1.1609        & 32.30 & 87.68              \\
  MAC      & $\times$              & 82.89	&79.46	&72.53	&55.86	&46.32	&40.50	&71.98                 & 1.4073	&1.3375&	1.3987       &38.64  & 147.80            \\
  BUTD     & $\checkmark$               & \textbf{90.01}	&82.02	&77.16	&60.95&	56.29&	42.29&	76.49                & 0.8905	 &1.2925	 &0.9501     & 34.95  & 177.55                   \\ 
  BAN     & $\checkmark$               & 89.81	&81.87	&77.58	&63.71&	55.67&	45.06&	76.74                & 0.8197	 &1.2417	 &0.8835        & 58.73 & 185.15                \\ 
  MCAN     & $\checkmark$              & 89.65	&81.65	&79.83	&63.16	&57.28	&43.71	&77.07                 & 0.8169	&1.2307	&0.8793      & 55.17 & 200.39                  \\ 
  D-VQA     & $\checkmark$               & 89.73	&82.12	&77.38	&63.99&	55.14&	43.20&	76.59                & 0.9167	 &1.2380	 &0.9627     & 37.79 & 179.29                   \\ \hline
  BLIP-2 & $\times$    & 88.13	& 81.92	&70.26	&58.58&	42.72&	28.34&	71.07                & 1.8790	 &1.3200	 &1.8186     & $\approx$4B & -                   \\ 
  Instruct-BLIP & $\times$    & 89.67	& 79.69	&76.96	&63.34&	59.72&45.68&	75.25                & 0.7994	 &1.2170	 &0.8627     & $\approx$4B & -                   \\ \hline
  \multicolumn{2}{l|}{$\star$\textsl{Remote sensing methods}}   & & & & & & & & & & \\
  RSVQA    & $\times$              & 82.43	&79.34	&70.68	&55.53	&42.45	&35.46	&70.70                & 1.7336	&1.3597	&1.6914    & 30.21 & 86.58               \\
  RSIVQA   & $\times$              & 85.32	& 80.44	& 75.01	& 56.63& 	51.55	& 39.25	& 73.71                & 1.7187	&1.3468	&1.6768     & 41.41 & 85.67               \\
\textbf{SOBA} (ours)& $\checkmark$   & 89.63 & \textbf{82.64} & \textbf{80.17} & \textbf{67.86} & \textbf{61.40} & \textbf{49.30} & \textbf{78.14} & \textbf{0.7856} & \textbf{1.1457} & \textbf{0.8391} & 40.46 & 185.69       \\ 
\end{tabular}}
\end{table*}

\section{Experiments}
\label{sec:5}
\noindent \textbf{Evaluation metrics.}
Following common settings \cite{yu2019deep},
we adopt the classification accuracy and root-mean-square error (RMSE) as evaluation metrics.
Especially, RMSE is used to evaluate counting tasks.
We use mean Union over Intersection (mIoU) to report semantic segmentation performance.
All experiments were performed under PyTorch framework using one RTX 3090 GPU.

\noindent \textbf{Experimental settings.}
For comparison,
we selected eight general (SAN \cite{yang2016stacked}, MAC \cite{mac}, BUTD \cite{anderson2018bottom}, BAN \cite{kim2018bilinear}, MCAN \cite{yu2019deep}, D-VQA \cite{wen2021debiased}, BLIP-2 \cite{blip2}, Instruct-BLIP \cite{dai2023instructblip}) and two remote sensing (RSVQA \cite{9088993}, RSIVQA \cite{zheng2021mutual}) VQA methods.
Because MCAN, BUTD, BAN, and D-VQA need semantic prompts, we adopt visual prompts from Semantic-FPN \cite{kirillov2019panoptic} fairly.
All VQA models were trained for 40k steps with a batch size of 16.
We set the two-layer LSTM with the hidden size of 384 and ResNet50 as default.
As for large vision-language models, BLIP-2 and Instruct-BLIP trained Q-Former following their original settings.
The vision encoder adopts ViT-g/14 and language decoder is FlanT5{\tiny XL}.
Following \cite{wang2021loveda},
Semantic-FPN was trained for 15k steps using the same batch size, generating visual prompts and semantic masks.
Segmentation augmentations include random flipping, rotation, scale jittering, and cropping for $512 \times 512$ patches.
We used Adam solver with $\beta_1=0.9$ and $\beta_2=0.999$.
The initial learning rate was set to $5e^{-5}$, and a `poly' schedule with a power of 0.9 was applied. 
The hidden size of the language and image features was $d_m = 384$.
The number of heads $M$ is set to 8, and the numbers of layers in self- and cross-attention modules are $N_E = N_D = 3$. 
We set $\alpha=1$ and $\gamma=0.5$ for ND loss.

\subsection{Comparative experiments}
\noindent \textbf{Main comparative results.}
Thanks to the diverse questions, EarthVQA can measure multiple perspectives of VQA models.
Table~\ref{tab:comp_rescuevqa} shows that all methods achieve high accuracies on basic judging questions.
The models with pixel-level visual prompts obtain higher accuracies, especially for the counting tasks.
This is because the semantic locations provide more spatial details, which benefits the object statistics.
Compared with advanced methods, SOBA achieves the best overall performances with similar or lower complexity.

\noindent \textbf{Object guided attention.}
OGA introduces object semantics into visual prompts and we compare it with related variants.
Table \ref{tab:oga} shows compared results for spatial, channel, and combined attentions, i.e, SA\cite{woo2018cbam}, SCSE\cite{roy2018recalibrating}, CBAM\cite{woo2018cbam}, SE\cite{hu2018squeeze}, GC\cite{cao2019gcnet}.
Channel attentions bring more stable improvements than spatial attentions. 
Because pseudo masks and visual prompts are concatenated dimensionally,
spatial attentions are hard to calibrate the subspaces of visual prompts and object masks.
Channel attentions enhance key object semantics and weaken uninterested background features.
Hence, our OGA abandoned spatial attention and achieved the best accuracies.

\begin{table}[!hbt]
  \centering
  \caption{Compared results with other attention mechanisms. `\texttt{C}' and `\texttt{S}' denote channel and spatial attention.}
  \label{tab:oga}
  \resizebox{0.85\linewidth}{!}{
  \begin{tabular}{l|c|cc}
  Object Guidance & Att. Type & $\uparrow$OA(\%)    & $\downarrow$OR   \\ \Xhline{1.0pt}
  Only Concat        & -    & 77.61      &   0.856      \\
  +SA            & \texttt{S}  & 77.72 &	0.861 \\
  +SCSE            & \texttt{C\&S}  & 77.89 & 0.854 \\
  +CBAM           & \texttt{C\&S}  & 77.95 & 0.857 \\
  +SE             & \texttt{C}    & 78.02 & 0.853 \\
  +GC             & \texttt{C}    & 78.03 & 0.847 \\ 
  +\textbf{OGA} (ours)      & \texttt{C}  & \textbf{78.14} & \textbf{0.839} \\
  \end{tabular}}
\end{table}

\noindent \textbf{One-way \textsl{vs.} bidirectional cross-attention.}
Existing transformer-based methods \cite{yu2019deep, cascante2022simvqa} utilize one-way (vanilla) attention to perform interactions, where
visual features are only treated as queries.
In contrast, we further gather enhanced visual features via the keywords (language features as queries), simulating the human process of finding visual cues.
As cross-attention consists of six transformer blocks, we compare the different combinations.   
Table~\ref{tab:oneway_vs_bca} shows that in one-way attention, querying visual features outperforms querying the language features.
This is because visual features are more informative, and their enhancement brings more improvements.
Bidirectional attention outperforms one-way structure due to more comprehensive interactions.

\begin{table}[!hbt]
  \centering
  \caption{Compared results between one-way (vanilla) and bidirectional cross-attention. `\texttt{V}' and `\texttt{L}' denote visual and language features, respectively.} \label{tab:oneway_vs_bca}
  \resizebox{0.85\linewidth}{!}{
  \begin{tabular}{l|l|cc}
  Cross-Attention          & Query   &$\uparrow$OA(\%)    & $\downarrow$OR  \\ \Xhline{1.0pt}
  \multirow{2}{*}{One-way (vanilla)} & \texttt{LLLLLL}     & 77.11                       & 0.977                         \\
                           & \texttt{VVVVVV}     & 77.53                       & 0.880                         \\ \hline
  \multirow{2}{*}{Bidirectional}    & \texttt{LLL-VVV}    & 77.57                       & 0.867                          \\
                                    & \texttt{VVV-LLL} & \textbf{78.14}        &\textbf{0.839}                         \\
  \end{tabular}}
  \end{table}

\begin{table}[!hbt]
  \centering
  \caption{Architecture ablation study}
  \label{tab:module_analysis}
  \resizebox{1.0\linewidth}{!}{
   \begin{tabular}{ccccc|cc}
   VSA       & BCA      & Promp.            & OGA & ND & $\uparrow$OA (\%)   & $\downarrow$OR     \\ \Xhline{1.0pt}
   $\checkmark$                    &                      &  &    &                        & 72.55          & 1.509       \\
                        & $\checkmark$                    &  &    &                        & 73.78          & 1.520                            \\
   $\checkmark$                    & $\checkmark$                    &                     &                         &                        & 74.91                          & 1.128                          \\
   $\checkmark$                    & $\checkmark$                    & $\checkmark$                    &                         &                        & 77.30              & 0.866\\
   $\checkmark$                    & $\checkmark$                    & $\checkmark$                    & $\checkmark$   &                        &  77.54                    & 0.859\\
   $\checkmark$                    & $\checkmark$                    & $\checkmark$                    & $\checkmark$   & $\checkmark$  & \textbf{78.14}      & \textbf{0.839}                     \\  
   \end{tabular}}
\end{table}

\subsection{Module analysis}

\noindent \textbf{Architecture of SOBA.}
SOBA was disassembled into five sub-modules: 1) VSA, 2) BCA, 3) semantic prompts, 4) OGA, and 5) ND loss.
Table~\ref{tab:module_analysis} shows that each module enhances the overall performance in distinct ways.
BCA produces a more significant improvement than VSA, and they complement each other (jointly obtaining OA=74.91\%).
OGA further improves the OA by explicitly adding the objectness semantics.
ND loss significantly boosts the counting performance from the aspect of optimization.
All modules are compatible with each other within the SOBA framework.

\noindent \textbf{Encoder variants.}
Table~\ref{tab:encoder_variants} shows the effects brought by segmentation networks with advanced CNN and Transformer encoders, \textsl{i.e.}, HRNet \cite{wang2020deep}, Swin Transformer \cite{liu2021swin}, Mix Transformer \cite{xie2021segformer}, ConvNeXt \cite{liu2022convnet}.
SOBA is compatible with the mainstream encoders and VQA performance is stable at a high level (OA$>$77.22\%).
Although MiT-B3 achieves lower segmentation accuracies than HR-W40, their features provide similar VQA performances.
As for similar segmentation architectures, larger encoders (Swin-S and ConvX-S) outperform better than smaller encoders (Swin-T and ConvX-T) in segmentation and VQA tasks.
With Wikipedia's
external knowledge, pretrained BERT-Base \cite{kenton2019bert} brings stable improvements.
With abundant computing power and time, larger encoders are recommended.

\begin{table}[!hbt]
  \centering
  \caption{Encoder variants analysis} \label{tab:encoder_variants}
  \resizebox{1\linewidth}{!}{
    \begin{tabular}{l|c|c|cc}
        Img Enc & Lan Enc & Param.(M)& $\uparrow$mIoU(\%)  & $\uparrow$OA(\%)         \\\Xhline{1.0pt}
        HR-W40  & LSTM &57.87& 57.31 & 77.92  \\
        MiT-B3  & LSTM &60.30& 56.44 & 77.43  \\ 
        Swin-T  & LSTM &43.86& 56.89 & 77.22  \\ 
        Swin-S  & LSTM &65.17& 57.44 & 78.01  \\ 
        ConvX-T & LSTM &44.16& 57.17 & 78.24  \\ 
        ConvX-S & LSTM &65.79& \textbf{57.34} & 78.43  \\\hline 
        Swin-T  & BERT-Base &153.42 & 56.89 & 77.63  \\
        Swin-S  & BERT-Base & 174.74& 57.44 & 78.23  \\
        ConvX-S  & BERT-Base & 175.36& \textbf{57.34} & \textbf{78.65}  \\ 
        \end{tabular}        }
\end{table}
\noindent \textbf{Bidirectional cross-attention variants.}
We explored BCA variants with different orders of query, \textsl{i.e.}, \texttt{V} and \texttt{L} were processed alternately, cascade, and parallel.
Table~\ref{tab:bca} shows that cascade structure \texttt{VVV-LLL} achieves the best accuracies.
\texttt{VVV} hierarchically aggregates language features to enhance visual features, and
\texttt{LLL} compresses the visual features to supplement language features.
Compared with \texttt{LLL}, first considering \texttt{VVV} retains the most information.
Hence, \texttt{VVV-LLL} represents the integration process from details to the whole, which conforms to human perception \cite{savage2019ai}.
Parallel structure obtains a sub-optimal accuracy, and frequent alternation of cross-attentions may lead to feature confusion.
\begin{table}[hbt]
  \begin{minipage}[t]{0.45\linewidth}
    \caption{BCA variants} \label{tab:bca}
    \centering
    \begin{tabular}{l|c}
       Query       & $\uparrow$OA(\%)      \\ \Xhline{1.0pt}
       \texttt{LV-LV-LV} & 77.51  \\
       \texttt{VL-VL-VL} & 77.58  \\ 
       \texttt{LLL-VVV}  & 77.57  \\
       \texttt{VVV-LLL} & \textbf{78.14} \\ 
       Parallel                   & 77.98 \\
       \end{tabular}
  \end{minipage}
  \hfill
  \begin{minipage}[t]{0.45\linewidth} 
    \centering
    \caption{Optim analysis} \label{tab:comp_optim}
    \resizebox{0.85\linewidth}{!}{
    \begin{tabular}{l|c}
      Optim.       & $\uparrow$OA(\%)     \\ \Xhline{1.0pt}
      CE & 77.54 \\
      DIW  & 77.38\\ 
      Focal  & 77.57	\\
      OHEM  & 77.85  \\ 
      SOM & 77.58	 \\
      \textbf{ND} & \textbf{78.14}               \\   
      \end{tabular}}
  \end{minipage}
\end{table}

\noindent \textbf{Optimization analysis.}
We compare ND loss with similar optimization algorithms designed to address the sample imbalance problem, including 1) dynamic inverse weighting (DIW) \cite{rajpurkar2017chexnet}, 2) Focal loss \cite{lin2017focal}, 3) online hard example mining (OHEM) \cite{shrivastava2016training}, 4) small object mining (SOM) \cite{factseg}.
In Table~\ref{tab:comp_optim}, Focal loss obtains better performance by adaptively balancing weights of easy and hard examples.
DIW failed to exceed the CE due to its extreme weighting strategies.
OHEM dynamic focuses on hard samples during the training, slightly improving OA (+0.31\%).
These optimization algorithms only focus on sample imbalances but are not sensitive to numerical distance.
They inherently cannot contribute to regression tasks.
In contrast, ND loss shows excellent performances on both classification and regression tasks.
\begin{figure}[hbt]
  \centering
  \includegraphics[width=1\linewidth]{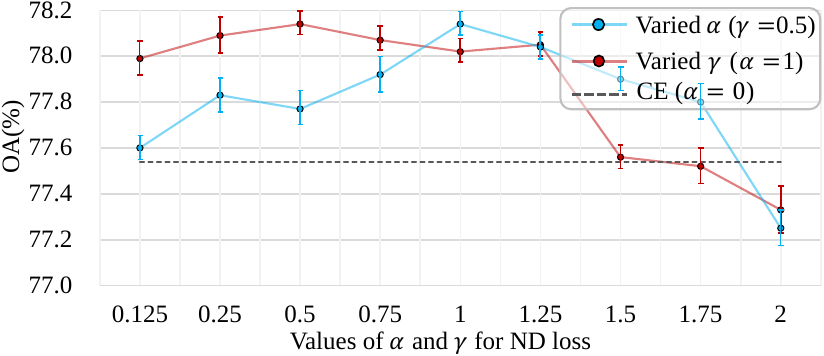}
  \caption{Experimental results with varied $\alpha$ and $\gamma$ for ND loss.   
  The optimal values range from 0.125 to 1.25 with wide ranges of hyperparameters selection.
  The mean values and standard deviations are reported after five runs.
  }
  \label{fig:alpha_gamma}
\end{figure}
\subsection{Hyperparameter analysis for ND loss}
As ND loss introduces two hyperparameters,
$\alpha$ controls overall penalty and $\gamma$ determines sensitivity to numerical differences.
In order to evaluate their effects on performances, we individually vary $\alpha$ and $\gamma$ from 0 to 2, and the results are reported in Figure~\ref{fig:alpha_gamma}.
Compared with CE loss, the additional difference penalty can bring stable gains. 
The suitable value of $\alpha$ ranges from 0.125 to 1.25 and reaches the highest OA at 1.
When $\alpha >$1.25, the performance drops because the large loss will bring instability during training.
When $\alpha$ is fixed at 1, 
the optional $\gamma$ also ranges from 0.125 to 1.25, and OA floats between 77.99\% and 78.14\%.
When $\gamma > 1$, the influence curve changes from concave to convex, resulting in a significant increase in difference penalties.
The model performance is not very sensitive to the hyperparameters introduced by ND loss, which reflects high fault tolerance and robustness.
Overall, our ND loss is superior to the CE baseline, with wide ranges of hyperparameter selection.

\begin{figure}[!hbt]
  \centering
  \subfigure[Classification loss ($\gamma_{0.5}$)]{
    \includegraphics[width=0.45\linewidth]{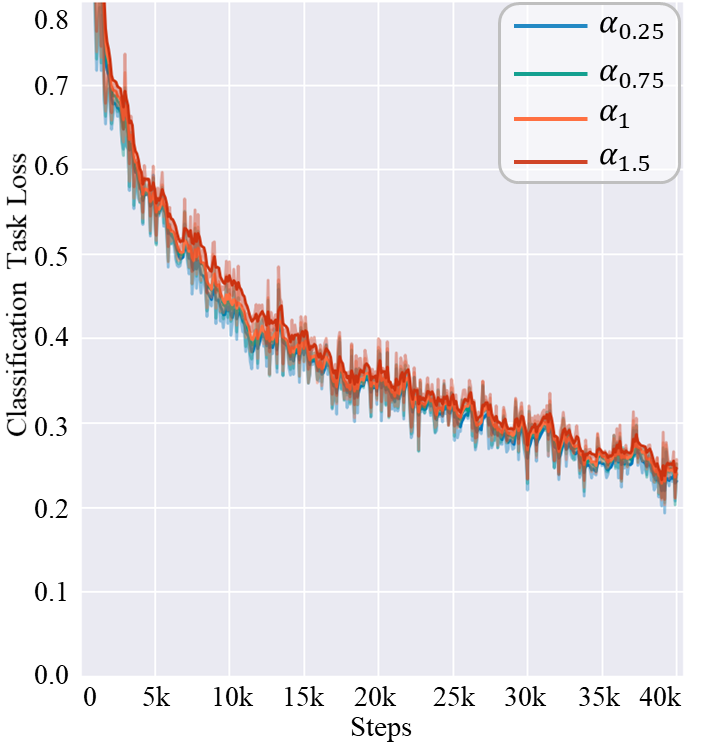}
    \label{fig:loss.sub1}}
  \subfigure[Regression loss ($\gamma_{0.5}$).]{
    \includegraphics[width=0.45\linewidth]{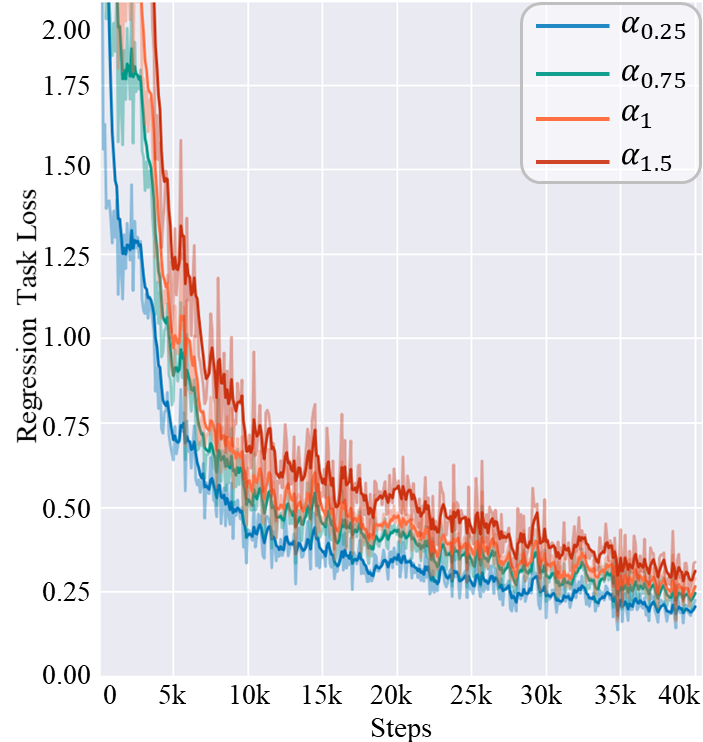}
    \label{fig:loss.sub2}}
  \subfigure[Classification loss ($\alpha_{1.0}$).]{
    \includegraphics[width=0.45\linewidth]{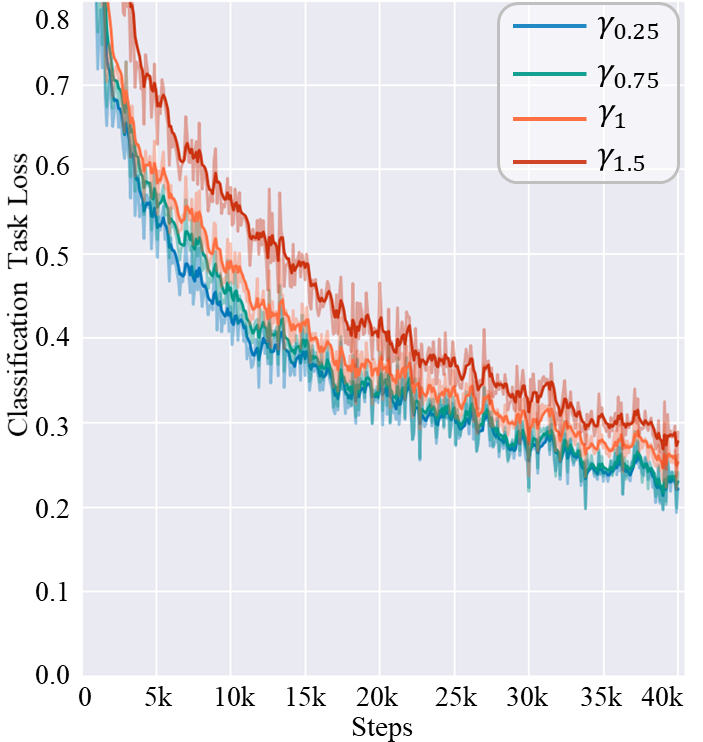}
    \label{fig:loss.sub3}}
  \subfigure[Regression loss ($\alpha_{1.0}$).]{
    \includegraphics[width=0.45\linewidth]{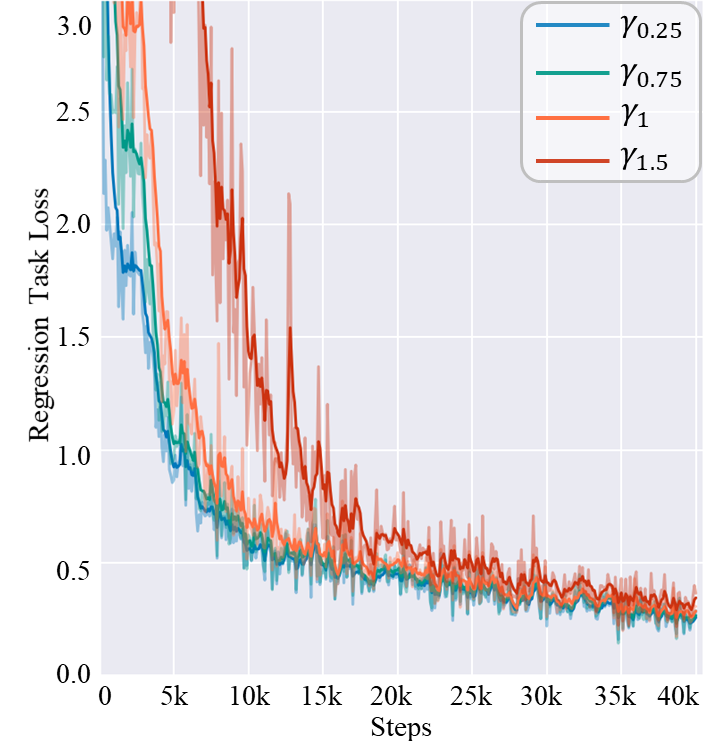}
    \label{fig:loss.sub4}}
  \caption{
	  The training losses of classification and regression tasks with different $\alpha$ and $\gamma$. 
    The changes of $\alpha$ and $\gamma$ mainly affect the regression task optimization.
    }
  \label{fig:loss}
\end{figure}

ND loss comprises two components, \textsl{i.e.}, the original classification loss and an
enhanced regression loss.
Figure~\ref{fig:loss} illustrates the effects of varying $\alpha$ and $\gamma$ on these two types of loss. 
It is evident that changes have little impact on classification optimization, as the difference penalty is only added to the regression loss. 
As the values of $\alpha$ and $\gamma$ increase, the regression losses become larger and more unstable.
However, as training progresses, the regression losses gradually stabilize and eventually converge.
Figure~\ref{fig:loss} shows that these two parameters control the numerical difference penalty in different ways.
This decomposition analysis of training loss can also provide references for tuning $\alpha$ and $\gamma$.

\subsection{Visualizations on bidirectional cross-attention}
\begin{figure}[!hbt]
  \centering
  \subfigure[How many \underline{intersections} are in this scene?]{
    \includegraphics[width=1\linewidth]{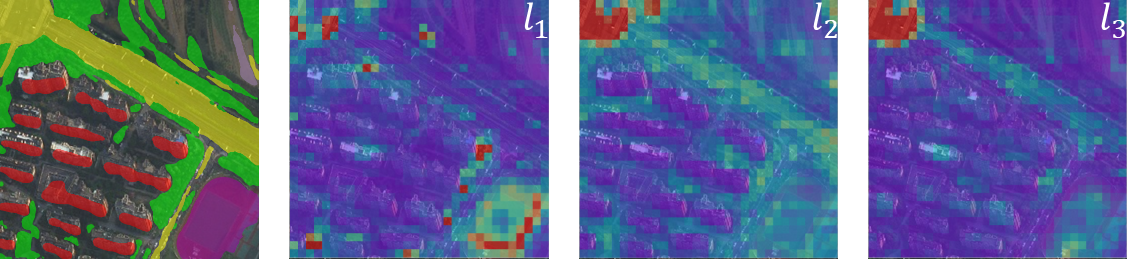}
    \label{fig:att.sub1}}
  \subfigure[What are the needs for the renovation of \underline{residents}?]{
    \includegraphics[width=1\linewidth]{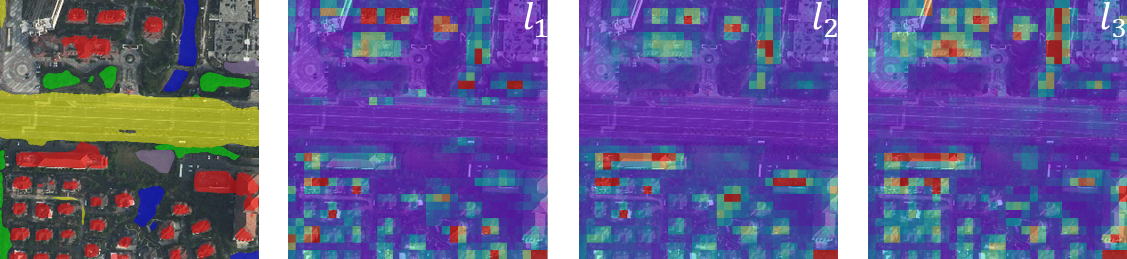}
    \label{fig:att.sub2}}
  \subfigure[What are the \underline{water} types in this scene?]{
    \includegraphics[width=1\linewidth]{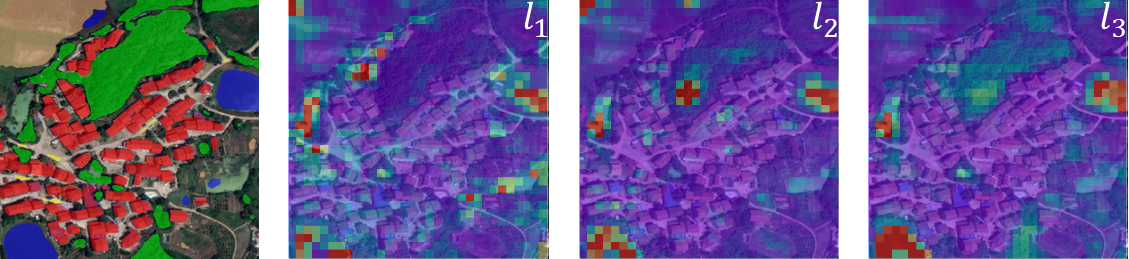}
    \label{fig:att.sub3}}
  \caption{
	  Visualization of attention maps in BCA with language features as queries. From left to right are the $l_1, l_2$ and $l_3$.
    Three examples are queried by different keywords: `\underline{intersections}', `\underline{residents}', and `\underline{water}'.
    }
  \label{fig:attention} 
\end{figure}

To analyze the mechanism of multi-modal feature interaction,
we visualize the attention maps in each layer of BCA according to different queries.
The question in Figure~\ref{fig:att.sub1} is `How many intersections are in this scene?', and `\underline{intersections}' is selected as a query word.
The first attention map shows some incorrect activations on the scattered roads and playground tracks.
However, as the layer deepens, BCA successfully reasons the right spatial relation for the key roads, and
the attention map focuses on the intersection in the upper left corner.
Similarly, Figure~\ref{fig:att.sub2} shows another example, which displays the process of gradually attending to the `\underline{residential}' area.
The third example shows a rural scene, and we select `\underline{water}' to query the visual features.
The attention map initially focuses on some trees and waters due to their similar spectral values.
Then the correct waters are enhanced, and uninterested trees are filtered out.

\section{Conclusion}
\label{sec:6}
To go beyond information extraction, we introduce the
VQA to remote sensing scene understanding, achieving
relational reasoning-based judging, counting, and situation analysis.
Based on the city planning needs, we designed a multi-modal
and multi-task VQA dataset named EarthVQA. Besides,
a two-stage semantic object awareness framework (SOBA) is
proposed to advance complex VQA tasks.
The extensive experiments
demonstrated the superiority of the proposed SOBA.
We hope the proposed
dataset and framework serve as a practical benchmark for
VQA in Earth observation scenarios.
Future work will explore the interactions between segmentation and VQA tasks.
\section{Acknowledgments}
This work was supported by National Natural Science Foundation of China under Grant Nos. 42325105, 42071350, and 42171336.
\bibliography{aaai24}

\end{document}